\documentclass{article}
\usepackage{wasysym}
\usepackage{arxiv}
\usepackage[T1]{fontenc}    % use 8-bit T1 fonts
\usepackage{hyperref}       % hyperlinks
\usepackage{url}            % simple URL typesetting
\usepackage{booktabs}       % professional-quality tables
\usepackage{amsfonts}       % blackboard math symbols
\usepackage{nicefrac}       % compact symbols for 1/2, etc.
\usepackage{microtype}      % microtypography
\usepackage{lipsum}
\usepackage{siunitx}
\usepackage{array}
\usepackage{tikz}
\usepackage{multicol}

\title{COVID-Twitter-BERT: A Natural Language Processing Model to Analyse COVID-19 Content on Twitter}
\author{
  Martin M\"uller \\
  Digital Epidemiology Lab \\
  EPFL \\
  Geneva, Switzerland \\
  \texttt{martin.muller@epfl.ch}
    \And
  Marcel Salath\'e \\
  Digital Epidemiology Lab \\
  EPFL \\
  Geneva, Switzerland \\
  \texttt{marcel.salathe@epfl.ch}
    \AND
  Per E Kummervold \\
  FISABIO-Public Health \\
  Vaccine Research Department \\
  Valencia, Spain \\
  \texttt{per@capia.no}
}

\begin{document}
\maketitle

\begin{abstract}
  In this work, we release COVID-Twitter-BERT (\textsc{CT-BERT}), a transformer-based model, pretrained on a large corpus of Twitter messages on the topic of COVID-19.
  Our model shows a \numrange{10}{30}\% marginal improvement compared to its base model, \textsc{BERT-Large}, on five different classification datasets.
  The largest improvements are on the target domain.
  Pretrained transformer models, such as \textsc{CT-BERT}, are trained on a specific target domain and can be used for a wide variety of natural language processing tasks, including classification, question-answering and chatbots.
  \textsc{CT-BERT} is optimised to be used on COVID-19 content, in particular from social media.
\end{abstract}

% keywords can be removed

\begin{multicols}{2}
\section{Introduction}
\label{sec:introduction}
Twitter has been a valuable source of news and a public medium for expression during the COVID-19 pandemic.
However, manually classifying, filtering and summarising the large amount of information available on COVID-19 on Twitter is impossible and has also been a challenging task to solve with tools from the field of machine learning and natural language processing (NLP).
To improve our understanding of Twitter messages related to COVID-19 content as well as the analysis of this content, we have therefore developed a model called COVID-Twitter-BERT (\textsc{CT-BERT})\footnote{\url{https://github.com/digitalepidemiologylab/covid-twitter-bert}}.

Transformer-based models have changed the landscape of NLP.
Models such as BERT, RoBERTa and ALBERT are all based on the same principle -- training bi-directional transformer models on huge unlabelled text corpuses~\cite{vaswani2017attention,devlin2018bert,liu2019roberta,lan2019albert}.
This process is done using methods such as mask language modelling (MLM), next sentence prediction (NSP) and sentence order prediction (SOP).
Different models vary slightly in how these methods are applied, but in general, all training is done in a fully unsupervised manner.
This process generates a general language model that is then used as input for a supervised finetuning for specific language processing tasks, such as classification, question-answering models, and chatbots.\par

Our model is based on the \textsc{BERT-Large} (English, uncased, whole word masking) model.
\textsc{BERT-Large} is trained mainly on raw text data from Wikipedia (3.5B words) and a free book corpus (0.8B words)~\cite{devlin2018bert}.
Whilst this is an impressive amount of text, it still contains little information about any specific subdomain.
To improve performance in subdomains, we have seen numerous transformer-based models trained on specialised corpuses.
  Some of the most popular ones are \textsc{BioBERT}~\cite{lee2020biobert} and \textsc{SciBERT}~\cite{beltagy2019scibert}.
These models are trained using the exact same unsupervised training techniques as the main models (MLM/NSP/SOP).
They can be trained from scratch, but this requires a very large corpus, so a more common approach is to start with the trained weights from a general model.
In this study, this process is called domain-specific pretraining.
When trained, such models can be used as replacements for general language models and be trained for downstream tasks.

\section{Method}
\label{sec:method}
The \textsc{CT-BERT} model is trained on a corpus of \num{160}M tweets about the coronavirus collected through the Crowdbreaks platform~\cite{muller2019crowdbreaks} during the period from January 12 to April 16, 2020.
Crowdbreaks uses the Twitter filter stream API to listen to a set of COVID-19-related keywords\footnote{wuhan, ncov, coronavirus, covid, sars-cov-2} in the English language.
Prior to training, the original corpus was cleaned for retweet tags.
Each tweet was pseudonymised by replacing all Twitter usernames with a common text token.
A similar procedure was performed on all URLs to web pages.
  We also replaced all unicode emoticons with textual ASCII representations (e.g.\ \texttt{:smile:} for \smiley) using the Python emoji library\footnote{\url{https://pypi.org/project/emoji/}}.
In the end, all retweets, duplicates and close duplicates were removed from the dataset, resulting in a final corpus of 22.5M tweets that comprise a total of 0.6B words.
The domain-specific pretraining dataset therefore consists of 1/7th the size of what is used for training the main base model.
Tweets were treated as individual documents and segmented into sentences using the spaCy library~\cite{honnibal2017spacy}.

All input sequences to the BERT models are converted to a set of tokens from a \num{30000}-word vocabulary.
As all Twitter messages are limited to 280 characters, this allows us to reduce the sequence length to 96 tokens, thereby increasing the training batch sizes to \num{1024} examples.
We use a dupe factor of 10 on the dataset, resulting in 285M training examples and 2.5M validation examples.
A constant learning rate of 2e-5, as recommended on the official BERT GitHub\footnote{\url{https://github.com/google-research/bert}} when doing domain-specific pretraining.

Loss and accuracy was calculated through the pretraining procedure.
For every \num{100000} training steps, we therefore save a checkpoint and finetune this towards a variety of downstream classification tasks.
Distributed training was performed using Tensorflow 2.2 on a TPU v3-8 (128GB of RAM) for \SI{120}{\hour}.
\subsection{Evaluation}
To assess the performance of our model on downstream classification tasks, we selected five independent training sets.
Three of them are publicly available datasets, and two are from internal projects not yet published.
All datasets consist of Twitter-related data.

\subsubsection{COVID-19 Category (CC)}
This dataset is a subsample of the data used for training \textsc{CT-BERT}, specifically for the period between January 12 and February 24, 2020.
Annotators on Amazon Turk (MTurk) were asked to categorise a given tweet text into either being a personal narrative (33.3\%) or news (66.7\%).
The annotation was performed using the Crowdbreaks platform~\cite{muller2019crowdbreaks}.

\subsubsection{Vaccine Sentiment (VS)}
This dataset contains a collection of measles- and vaccination-related US-geolocated tweets collected between March 2, 2011 and October 9, 2016.
The dataset was first used by Pananos et al.~\cite{pananos2017critical}, but a modified version from M\"uller et al.~\cite{muller2019crowdbreaks} was used here.
The dataset contains three classes: positive (towards vaccinations) (51.9\%), negative (7.1\%) and neutral/others (41.0\%).
The neutral category was used for tweets which are either irrelevant or ambiguous. Annotation was performed on MTurk.

\subsubsection{Maternal Vaccine Stance (MVS)}
The dataset is from a so far unpublished project related to the stance towards the use of maternal vaccines.
Experts in the field annotated the data into four categories: neutral (41.0\%), discouraging (25.3\%), promotional (43.9\%) and ambiguous (14.3\%).
Each tweet was annotated threefold, and disagreement amongst the experts was resolved in each case by using a common scoring criterion.

\subsubsection{Twitter Sentiment SemEval (SE)}
This is an open dataset from SemEval-2016 Task 4: Sentiment Analysis in Twitter~\cite{nakov2019semeval}.
In particular, we used the dataset for subtask A, a dataset annotated fivefold into three categories: negative (15.7\%), neutral (45.9\%) and positive (38.4\%).
We make a small adjustment to this dataset by fully anonymising links and usernames.

\subsubsection{Stanford Sentiment Treebank 2 (SST-2)}
SST-2 is a public dataset consisting of binary sentiment labels, negative (44.3\%) and positive (55.7\%), within sentences~\cite{socher2013recursive}.
Sentences were extracted from a dataset of movie reviews~\cite{pang2005seeing} and did not originate from Twitter, making SST-2 our only non-Twitter dataset.

The dataset split size is predefined for the SST-2 and SE datasets.
For the SST-2 dataset, the test dataset is not released.
For the other datasets, we aimed at a split of around 50\%-30\% between the training and development sets, leaving a test set of 20\% which was not used in this work.
Our intention was not to optimise the finetuned models but to thoroughly evaluate the performance of the domain-specific \textsc{CT-BERT}-model.
We experimented with different numbers of epochs for each training dataset for \textsc{BERT-Large} (i.e.\ checkpoint 0 of \textsc{CT-BERT}) and selected the optimal one.
We then used this number in subsequent experiments on the respective dataset.
We ended with three epochs for SST-2, CC and SE, five epochs for VC and 10 epochs for MVC, all with a learning rate of 2e-05.
The number of epochs was dependent on both the size and balance of the categories.
Larger and unbalanced sets require more epochs.

\end{multicols}
\begin{table}
  \def\arraystretch{1.6} % vertical stretch factor
  \centering
  \begin{tabular}{lcrrrp{10mm}}
  \toprule
    Dataset                               & Classes & Train       & Dev    & \multicolumn{1}{c}{Labels} \\
    \midrule
    COVID-19 Category (CC)                & 2       & \num{3094}  & \num{1031} &
    \begin{tikzpicture}[baseline=-10]
      \draw (0,0) rectangle (1.67,-0.5) node[midway] {Personal};
      \draw (1.67,0) rectangle (5,-0.5) node[midway] {News} ;
    \end{tikzpicture}   \\
    Vaccine Sentiment (VC)                & 3       & \num{5000}  & \num{3000} &
    \begin{tikzpicture}[baseline=-10]
      \draw (0,0) rectangle (0.356,-0.5) node[midway] {N};
      \draw (0.356,0) rectangle (2.952,-0.5) node[midway] {Neutral} ;
      \draw (2.952,0) rectangle (5,-0.5) node[midway] {Positive} ;
    \end{tikzpicture}   \\
    Maternal Vaccine Stance (MVS)         & 4       & \num{1361}  & \num{817}  &
    \begin{tikzpicture}[baseline=-10]
      \draw (0,0) rectangle (1.267,-0.5) node[midway] {Disc};
      \draw (1.267,0) rectangle (1.982,-0.5) node[midway] {A} ;
      \draw (1.982,0) rectangle (2.267,-0.5) node[midway] {N} ;
      \draw (2.267,0) rectangle (5,-0.5) node[midway] {Promotional} ;
    \end{tikzpicture}   \\
    Stanford Sentiment Treebank 2 (SST-2) & 2       & \num{67349} & \num{872}  &
    \begin{tikzpicture}[baseline=-10]
      \draw (0,0) rectangle (2.214,-0.5) node[midway] {Negative};
      \draw (2.214,0) rectangle (5,-0.5) node[midway] {Positive} ;
    \end{tikzpicture}   \\
    Twitter Sentiment SemEval (SE)        & 3       & \num{6000}  & \num{817}  &
    \begin{tikzpicture}[baseline=-10]
      \draw (0,0) rectangle (0.783,-0.5) node[midway] {Neg};
      \draw (0.783,0) rectangle (3.080,-0.5) node[midway] {Neutral} ;
      \draw (3.080,0) rectangle (5,-0.5) node[midway] {Positive} ;
    \end{tikzpicture}   \\
    \bottomrule \\
  \end{tabular}
  \caption{
    Overview of the evaluation datasets.
    All five evaluation datasets are multi-class datasets with sometimes strong label imbalance, visualised by the proportional bar width in the label column.
    N and Neg stand for negative; Disc and A stand for discouraging and ambiguous, respectively.
  }
  \label{tab:tab1}
\end{table}
\begin{multicols}{2}

\section{Results}
\label{sec:results}

\subsection{Domain-sepcific pretraining}
\label{sec:domain_specific_pretraining}
\begin{figure}
  \centering
  \includegraphics[]{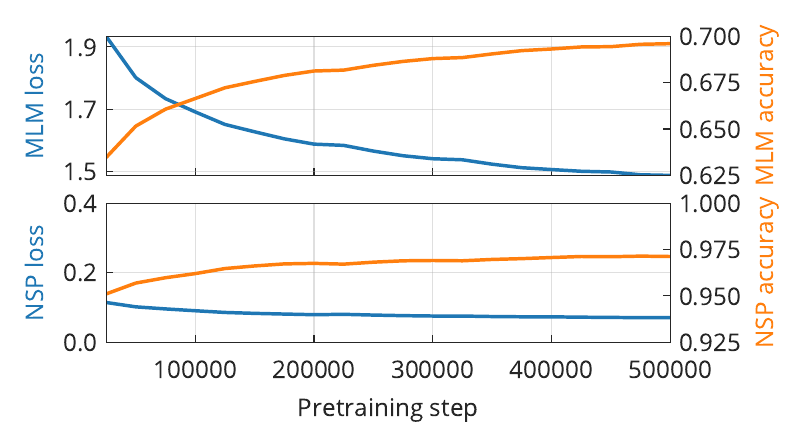}
  \caption{
    Evaluation metrics for the domain-specific pretraining of \textsc{CT-BERT}.
    Shown are the loss and accuracy of masked language modelling (MLM) and next sentence prediction (NSP) tasks.
  }
  \label{fig:fig1}
\end{figure}

Figure~\ref{fig:fig1} shows the progress of pretraining \textsc{CT-BERT} at intervals of 25k training steps and the evaluation of 1k steps on a held-out validation dataset.
All metrics considered improve throughout the training process.
The improvement on the MLM loss task is most notable and yields a final value of 1.48.
The NSP task improves only marginally, as it already performs very well initially.
Training was stopped at \num{500000}, an equivalent of \num{512}M training examples, which we consider as our final model.
This corresponds to roughly 1.8 training epochs.
All metrics for the MLM and NLM tasks improve steadily throughout training.
However, using loss/metrics for these tasks to evaluate the correct time to stop training is difficult.

\end{multicols}
\begin{multicols}{2}

\subsection{Evaluation on classification datasets}
\label{sec:evaluation_classification}

To assess the performance of our model properly, we compared the mean F1 score of \textsc{CT-BERT} with that of \textsc{BERT-Large} on five different classification datasets.
We adapted the number of training epochs for each dataset according to its size in order to have a similar number of training steps for each dataset.
  Our final model shows higher performance on all datasets (a mean F1 score of \num{0.833}) compared with \textsc{BERT-Large} (a mean F1 score of \num{0.802}).
As the initial performance varies widely across datasets, we compute the relative improvement in marginal performance ($\Delta$MP) for each dataset.
$\Delta$MP is calculated as follows:

$$ \Delta\text{MP} = \frac{F_{1, \text{ \textsc{BERT-Large}}} - F_{1, \text{ \textsc{CT-BERT}}}}{1- F_{1, \text{ \textsc{BERT-Large}}}} $$

From this metric, we can observe the largest improvement of our model on the COVID-19-specific dataset (CC), with a $\Delta$MP value of 25.88\%.
The marginal improvement is also high on the Twitter datasets related to vaccine sentiment (MVS).
Our model likewise shows some improvements on the SST-2 and SemEval datasets, but to a smaller extent.

\end{multicols}
\begin{table}
  \centering
  \begin{tabular}{lrrr}
    \toprule
    Dataset  & \textsc{BERT-Large} & \textsc{CT-BERT} & $\Delta\text{MP}$  \\
    \midrule
    COVID-19 Category (CC)                    & 0.931 & 0.949 & 25.88\%  \\
    Vaccine Sentiment (VC)                    & 0.824 & 0.869 & 25.27\%  \\
    Maternal Vaccine Stance (MVS)             & 0.696 & 0.748 & 17.07\%  \\
    Stanford Sentiment Treebank 2 (SST-2)     & 0.937 & 0.944 & 10.67\%  \\
    Twitter Sentiment SemEval (SE)            & 0.620 & 0.654 & 8.97\%   \\
    \midrule
    Average &  0.802 & 0.833 & 17.57\% \\
    \bottomrule \\
  \end{tabular}
  \caption{
    Comparison of the final model performance with \textsc{BERT-Large}.
    \textsc{CT-BERT} shows improvements on all datasets.
    The marginal improvement is the highest on the COVID-19-related dataset (CC) and lowest on the SST-2 and SemEval datasets.
  }
  \label{tab:tab2}
\end{table}
\begin{multicols}{2}

\subsection{Evaluation on intermediary pretraining checkpoints}
\label{sec:evaluation_checkpoints}
\begin{figure}[h!]
  \centering
  \includegraphics[]{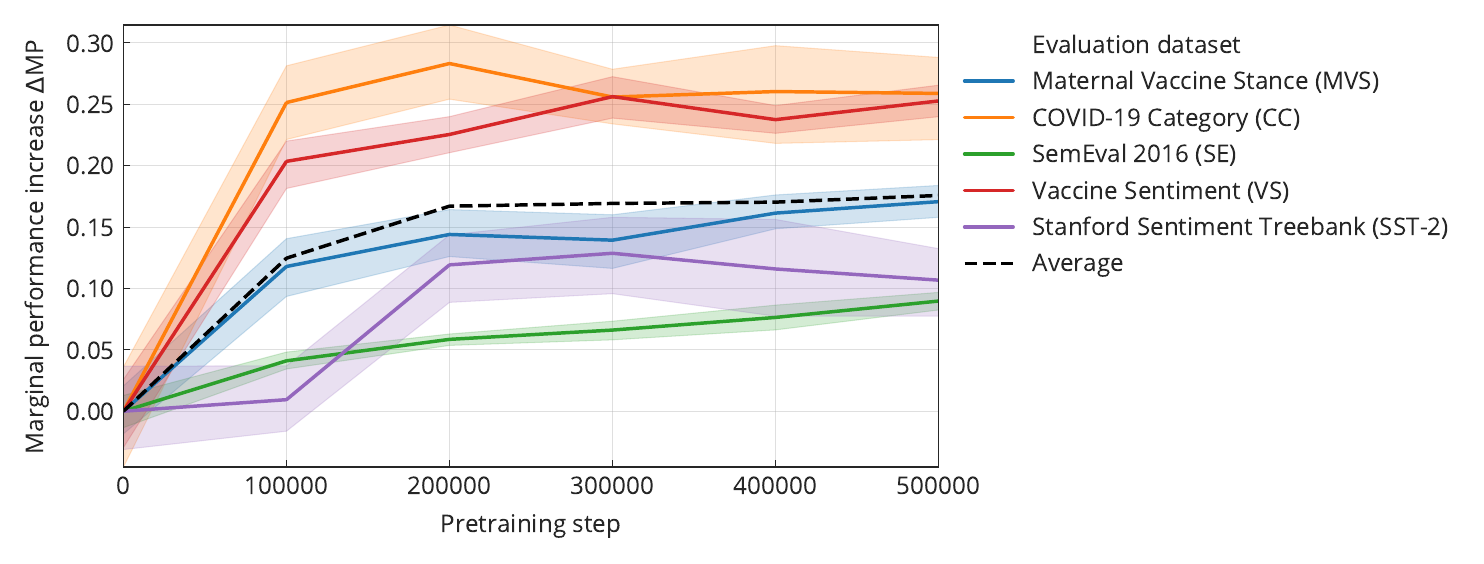}
  \caption{
    Marginal performance increase in the F1 score ($\Delta$MP) on finetuning on various classification tasks at increasing steps of pretraining.
    Zero on the x-axis corresponds to the base model, which is \textsc{BERT-Large} in this case.
    Our model improves on all evaluated datasets, with the biggest relative improvement being in the COVID-19 category dataset.
    The bands show the standard error of the mean (SEM) out of 10 repeats.
  }
  \label{fig:fig2}
\end{figure}

So far, we have seen improvements in the final \textsc{CT-BERT} model on all evaluated datasets.
  To understand whether the observed decrease in loss during pretraining linearly translates into performance on downstream classification tasks, we evaluated \textsc{CT-BERT} on five intermediary versions (checkpoints) of the model and on the zero checkpoint, which corresponds to the original \textsc{BERT-Large} model.
At each intermediary checkpoint, 10 repeated training runs (finetunings) for each of the five datasets were performed, and the mean F1 score was recorded.
Figure~\ref{fig:fig2} shows the marginal performance increase ($\Delta$MP) at specific pretraining steps.
Our experiments show that downstream performance increases fast up to step \num{200}k in the pretraining and only demonstrates marginal improvement afterwards.
The loss curve, on the other hand, shows a gradual increase even after step \num{200}k.
We also note that for the COVID-19-related dataset, most of the marginal improvement occurred after \num{100}k pretraining steps.
SST-2, the only non-Twitter dataset, improves much more slowly and reaches its final performance only after \num{200}k pretraining steps.

Amongst runs on the same model and dataset, some degree of variance in performance was observed.
This variance is mostly driven by runs with a particularly low performance.
We observe that the variance is dataset dependent, but it does not increase throughout different pretraining checkpoints and is comparable to the variance observed on \textsc{BERT-Large} (pretraining step zero).
The most stable training seems to be on the SemEval training set, and the least stable one is on SST-2, but most of this difference is within the error margins.

\end{multicols}
\begin{multicols}{2}

\section{Discussion}
\label{sec:discussion}
The most accurate way to evaluate the performance of a domain-specific model is to apply it on specific downstream tasks.
\textsc{CT-BERT} is evaluated on five different Twitter-based datasets.
Compared to \textsc{BERT-Large}, it improves significantly on all datasets.
However, the improvement is largest in datasets related to health, particularly in datasets related to COVID-19.
We therefore expect \textsc{CT-BERT} to perform similarly well on other classification problems on COVID-19-related data sources, but particularly on text derived from social media platforms.

Whilst it is expected that the benefit of using \textsc{CT-BERT} instead of \textsc{BERT-Large} is greatest when working with Twitter COVID-19 text, it is reasonable to expect some performance gains even when working with general Twitter messages (SemEval dataset) or with a non-Twitter dataset (SST-2).

Our results show that the MLM and NSP metrics during the pretraining align to some degree with downstream performance on classification tasks.
However, compared with COVID-19 or health-related content, out-of-domain text might require longer pretraining to achieve a similar performance boost.

Whilst we have observed an improvement in performance on classification tasks, we did not test our model on other natural language understanding tasks.
Furthermore, at the time of this paper’s writing, we only had access to one COVID-19-related dataset.
The general performance of our model might be improved further by considering pretraining under different hyperparameters, particularly modifications to the learning rate schedules, training batch sizes and optimisers.
Future work might include evaluation on other datasets and the inclusion of more recent training data.

The best way to evaluate pretrained transformer models is to finetune them on downstream tasks.
Finetuning a classifier on a pre-trained model is considered computationally cheap.
The training time is usually done in an hour or two on a GPU.
Using this method for evaluation is more expensive, as it requires evaluating multiple checkpoints to monitor improvement and on several varied datasets to show robustness.
As finetuning results vary between each run, each experiment must be performed multiple times when the goal is to study the pretrained model.
In this case, we repeated the training for six checkpoints, 10 runs for each checkpoint on all the five datasets.
A total of 300 evaluation runs were performed.
The computational cost for evaluation is therefore on par with the pretraining.
Large and reliable training and validation sets make this task easier, as the number of repetitions can be reduced.

All the tests are done on categorisation tasks, as this task is easier in terms of both data access and evaluation.
However, transformer-based models can be used for a wide range of tasks, such as named entity recognition and question answering.
It is expected that \textsc{CT-BERT} can also be used for these kinds of tasks within our target domain.

Our primary goal in this work was to obtain stable results on the finetuning in order to evaluate the pre-trained model, not to necessarily optimise the finetuning.
The number of finetuning epochs and the learning rate are, for instance, have been optimised for \textsc{BERT-Large}, not for \textsc{CT-BERT}.
This means that there is still great room for optimisation on the downstream task.

\section{Data Availability}
The model, code and public datasets are available in our GitHub repository: \url{https://github.com/digitalepidemiologylab/covid-twitter-bert}.

\section{Funding}
PK received funding from the European Commission for the call H2020-MSCA-IF-2017 and the funding scheme MSCA-IF-EF-ST for the VACMA project (grant agreement ID: 797876).

MM and MS received funding through the Versatile Emerging infectious disease Observatory grant as a part of the European Commission’s Horizon 2020 framework programme (grant agreement ID: 874735).

The research was supported with Cloud TPUs from Google's TensorFlow Research Cloud and Google Cloud credits in the context of COVID-19-related research.

\section{Conflicts of Interest}
The authors have no conflicts of interest to declare.

\end{multicols}

\bibliographystyle{unsrt}
\bibliography{refs}
\end{document}